\documentclass[10pt,twocolumn,letterpaper]{article}

\usepackage[pagenumbers]{cvpr} % To force page numbers, e.g. for an arXiv version

\usepackage{tabularx}
\usepackage{array}
\usepackage[table]{xcolor}

\definecolor{cvprblue}{rgb}{0.21,0.49,0.74}
\usepackage[pagebackref,breaklinks,colorlinks,allcolors=cvprblue]{hyperref}

\def\confName{CVPR}
\def\confYear{2027}

\title{AlignedHPO: Efficient Surrogate-Guided Loss Hyperparameter Optimization}

\author{Ivan Karpukhin\\
Sber AI Lab\\
{\tt\small iakarpukhin@sberbank.ru}
\and
Andrey Savchenko\\
Sber AI Lab\\
HSE University \\
ISP RAS Research Center for Trusted Artificial Intelligence\\
}

\begin{document}
\maketitle
\begin{abstract}
Modern deep learning losses involve multiple hyperparameters, such as weights, scales, and margins, typically tuned manually or through expensive black-box optimization despite their differentiability.

We propose AlignedHPO, an efficient framework for tuning differentiable loss hyperparameters using gradient alignment and surrogate validation objectives. Our method employs Jacobian-vector products to reduce hypergradient computation time and memory, while differentiable metric surrogates guide optimization toward task-specific evaluation criteria. We address gradient magnitude ambiguity and stochastic instability through normalized alignment and gradient smoothing. Experiments with generalized Focal Loss and ArcFace demonstrate effective adaptation across long-tailed classification and open-set face recognition. AlignedHPO achieves competitive or improved performance compared to implicit differentiation, with nearly 12× faster hypergradient computation. These results highlight the potential of efficient surrogate-guided loss hyperparameter optimization.

% TODO: not only forward pass, but also validation FW/BW step.
\end{abstract}

\section{Introduction}

%\the\linewidth
%237.13594pt

\begin{figure}[t]
    \centering

    \begin{subfigure}{\columnwidth}
        \centering
        \includegraphics[width=\linewidth]{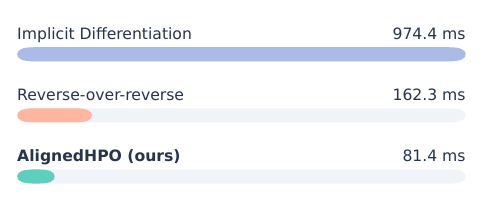}
        \caption{Hypergradient computation time.}
        \label{fig:hypergradient_time}
    \end{subfigure}

    \vspace{1mm}

    \begin{subfigure}{\columnwidth}
        \centering
        \includegraphics[width=\linewidth]{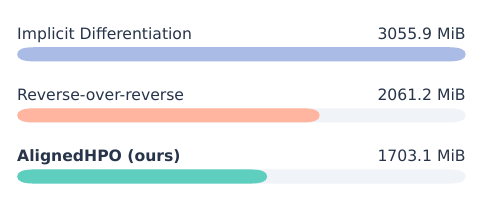}
        \caption{Peak GPU memory consumption.}
        \label{fig:hypergradient_memory}
    \end{subfigure}

    \caption{
        Computational efficiency of hypergradient estimation
        for an ArcFace model with a ResNet-50 backbone
        (batch size 32).
        We compare implicit differentiation using a
        five-step Neumann approximation,
        conventional reverse-over-reverse differentiation,
        and our forward-over-reverse implementation
        (AlignedHPO).
        %(a) Median computation time per hypergradient.
        %(b) Incremental peak GPU memory consumption.
        Lower values are better.
    }
    \label{fig:hypergradient_benchmark}
\end{figure}

Loss function design plays a central role in deep
learning across applications such as computer vision,
natural language processing, and reinforcement learning.
Modern loss functions often involve multiple
hyperparameters, including weighting coefficients,
scales, margins, and focusing factors, which influence
training dynamics and final model performance.
Despite their importance, these hyperparameters are
typically selected through manual experimentation or
expensive black-box optimization, often requiring
numerous training runs.

Beyond computational cost, hyperparameter tuning
complicates the development and evaluation of new
training objectives. Performance improvements may
depend as much on hyperparameter selection as on the
loss formulation itself, while competing methods are
rarely tuned with equal care. Consequently, suboptimal
hyperparameter choices can bias experimental comparisons
and undermine the reliability of empirical conclusions.

Many loss hyperparameters, however, are differentiable,
enabling gradient-based optimization through their
influence on the training trajectory. This principle
underlies hypergradient and bilevel optimization
methods, which have been applied to optimization
parameters, loss weights, and learned training
objectives. Nevertheless, efficiently tuning general
loss hyperparameters toward task-specific evaluation
metrics remains challenging.

Existing approaches to automatic loss optimization
typically rely on differentiation through the training
trajectory or implicit differentiation. The former
can introduce substantial computational and memory
overhead when unrolling multiple optimization steps.
The latter avoids unrolling but generally assumes an
approximately stationary training objective and
requires expensive inverse-Hessian--vector product
approximations. These limitations make both approaches
costly for online hyperparameter optimization in
large neural networks.

Another challenge lies in the choice of validation
objective. Evaluation metrics, such as classification
accuracy and ROC AUC, are often non-differentiable,
while conventional validation losses may not accurately
reflect the target metric. Differentiable metric
surrogates offer a way to bridge this gap, allowing
loss hyperparameters to be optimized toward the
desired evaluation criterion.

In this work, we propose AlignedHPO, an efficient
framework for surrogate-guided optimization of
differentiable loss hyperparameters during training.
Our method combines local gradient alignment with
forward-mode automatic differentiation, using
Jacobian--vector products to efficiently estimate
hypergradients without unrolling training trajectories
or approximating inverse Hessians. We further address
limitations of direct gradient alignment, including
undesirable gradient scaling and stochastic instability,
through normalized similarity objectives and
historical gradient smoothing.

We evaluate AlignedHPO on generalized Focal Loss
and ArcFace objectives, jointly optimizing up to
hundreds of hyperparameters. Experiments on long-tailed
classification demonstrate improvements over fixed
loss configurations and competitive performance
with implicit differentiation at substantially lower
computational cost. Open-set face recognition
experiments further demonstrate dataset-specific
adaptation while highlighting the challenges of
generalization across different validation distributions.

Our main contributions are summarized as follows:
\begin{enumerate}
    \item We introduce a framework for online
    optimization of differentiable loss hyperparameters,
    including nonlinear parameterizations, guided by
    surrogate validation objectives that approximate
    task-specific evaluation metrics.

    \item We develop an efficient hypergradient
    computation procedure based on Jacobian--vector
    products, substantially reducing runtime and
    memory consumption compared to conventional
    reverse-over-reverse and implicit differentiation.

    \item We analyze challenges of gradient alignment,
    including sensitivity to gradient magnitude and
    stochastic instability, and introduce normalized
    alignment and gradient smoothing to improve
    optimization robustness.

    \item We demonstrate effective loss hyperparameter
    adaptation on long-tailed classification and
    open-set face recognition, investigating both
    computational efficiency and generalization
    across evaluation distributions.
\end{enumerate}
% \input{sec/0_abstract}    
% \input{sec/1_intro}
% \input{sec/2_formatting}
% \input{sec/3_finalcopy}

% Preamble:
% \usepackage{booktabs}
% \usepackage{amssymb}
% \usepackage[table]{xcolor}
% \usepackage{graphicx}

% Preamble:
% \usepackage{booktabs,amssymb}
% \usepackage[table]{xcolor}

% Preamble:
% \usepackage{booktabs,tabularx,array,amssymb}
% \usepackage[table]{xcolor}

\definecolor{tableblue}{RGB}{230,241,249}

\begin{table*}[t]
\centering
\caption{Comparison of differentiable loss optimization
methods. We summarize their applications, support
for nonlinear loss parameters, tuning strategies,
target objectives, and computational complexity.
Complexities indicate the number of differentiation
operations rather than directly comparable runtimes.}
\label{tab:related_comparison}

\begingroup
\fontsize{10}{11.5}\selectfont
\setlength{\tabcolsep}{4pt}
\renewcommand{\arraystretch}{1.08}

\begin{tabularx}{\textwidth}{
@{}l l c l l X@{}
}
\toprule
\textbf{Method} &
\textbf{Application} &
\textbf{Nonlinear param.} &
\textbf{Tuning} &
\textbf{Target objective} &
\textbf{Complexity} \\
\midrule

OL-AUX~\cite{lin2019adaptive}
 & AUX & --
 & Gradient alignment
 & Main loss
 & $O(K)$ BP \\

Auto-$\lambda$~\cite{liu2022autolambda}
 & MTL & --
 & Bilevel (FD)
 & Val. task loss
 & $O(1)$ FP/BP \\

AMAL~\cite{sivasubramanian2023amal}
 & AUX & --
 & One-step bilevel
 & Val. task loss
 & $O(1)$ BP \\

Meta-Weight-Net~\cite{shu2019metaweight}
 & Design/HPO & --
 & One-step bilevel
 & Cross-Ent.
 & $O(1)$ BP \\

L2T-DLF~\cite{wu2018learning}
 & Design/HPO & $\checkmark$
 & Unrolled bilevel
 & Expected metric
 & $O(T)$ HVP + mixed \\

ML3~\cite{bechtle2021ml3}
 & Design/HPO & $\checkmark$
 & Unrolled bilevel
 & Meta-task loss
 & $O(T)$ BP/HVP \\

AuxiLearn~\cite{navon2021auxilearn}
 & AUX & $\checkmark^{*}$
 & Implicit diff.
 & Val. main loss
 & $O(Q)$ HVP \\

AutoBalance~\cite{li2021autobalance}
 & Design/HPO & $\checkmark^{\dagger}$
 & Implicit diff.
 & Balanced loss
 & $O(Q)$ HVP \\

\midrule
\rowcolor{tableblue}
\textbf{AlignedHPO}
 & \textbf{Design/HPO} & $\checkmark$
 & \textbf{Gradient alignment}
 & \textbf{Val. surrogate}
 & $\boldsymbol{O(1)}$ \textbf{JVP} \\

\bottomrule
\end{tabularx}
\endgroup

\vspace{3pt}
{\fontsize{8}{9.5}\selectfont
\parbox{\textwidth}{
$K$: auxiliary losses; $T$: unrolled steps;
$Q$: inverse-Hessian iterations.
FP/BP/JVP/HVP: forward/backward,
Jacobian-vector, and Hessian-vector products.
Target objective denotes the criterion guiding
loss-parameter updates, not the evaluation metric.
$^{*}$Nonlinear combinations of predefined
auxiliary losses.
$^{\dagger}$Class-specific parameters of a
cross-entropy-based loss.
Costs exclude standard training and validation-gradient
computation; constant factors and loss-level operations
are omitted.
}}
\end{table*}

\section{Related Work}
\label{sec:related_work}

\paragraph{Gradient-based hyperparameter optimization.}
Gradient-based HPO exploits the dependence of training
on hyperparameters. Early methods differentiated through
complete trajectories using reversible learning
\cite{maclaurin2015reversible}, followed by forward-
and reverse-mode differentiation
\cite{franceschi2017forwardreverse} and truncated
backpropagation~\cite{shaban2019truncated}.
Implicit differentiation instead estimates hypergradients
at approximately stationary solutions, requiring
inverse-Hessian--vector product approximations
\cite{lorraine2020millions}. Recent stochastic formulations
relax stationarity assumptions~\cite{kim2025stochastic}.
However, trajectory differentiation incurs substantial
computational and memory costs, while implicit methods
require expensive Hessian-related computations.
We focus on efficient online optimization without
long unrolling or stationarity assumptions.

\paragraph{Adaptive weighting and auxiliary learning.}
Related methods adapt the contributions of training
examples, tasks, and auxiliary losses. Gradient similarity
prevents harmful auxiliary updates~\cite{du2018adapting},
while OL-AUX estimates auxiliary gradient contributions
\cite{lin2019adaptive}. Validation-based meta-gradients
enable example reweighting~\cite{ren2018reweight},
extended to nonlinear weighting by Meta-Weight-Net
\cite{shu2019metaweight}. Auto-$\lambda$ learns task
weights through bilevel optimization
\cite{liu2022autolambda}, while AMAL adapts instance-level
loss mixtures~\cite{sivasubramanian2023amal}. 
Gradient-based loss balancing has also been explored
in GradNorm~\cite{chen2018gradnorm}, while
PCGrad~\cite{yu2020gradient} uses gradient
projections to mitigate conflicting updates.
These approaches primarily optimize example or task
weights rather than internal loss hyperparameters.

\paragraph{Automatic loss design.}
L2T-DLF learns training losses using a teacher network
and trajectory differentiation~\cite{wu2018learning}.
AuxiLearn uses implicit differentiation to learn nonlinear
combinations of auxiliary losses and tasks
\cite{navon2021auxilearn}, while AutoBalance optimizes
class- and group-dependent loss parameters for imbalanced
classification~\cite{li2021autobalance}.
These methods support nonlinear objectives but incur
costly trajectory differentiation or iterative
inverse-Hessian approximations. In contrast, we optimize
general differentiable loss hyperparameters without
unrolling or Hessian estimation, demonstrating our method
on Focal Loss and ArcFace
\cite{lin2017focal,deng2019arcface}.

\paragraph{Efficient gradient alignment and hypergradients.}
Reverse-mode differentiation computes one-step
hypergradients without explicitly constructing
mixed-derivative matrices
\cite{franceschi2017forwardreverse,ren2018reweight}.
More closely related, GraP~\cite{karpukhin2026grap}
optimizes linear weights of composite pretraining losses
through gradient alignment in the shared embedding space,
using normalization to address gradient-scale ambiguity.
Our method extends alignment to nonlinear loss
hyperparameters and employs forward-over-reverse
Jacobian--vector products for efficient computation.
We additionally introduce historical gradient smoothing
to improve stability under stochastic training.

\paragraph{Surrogate validation objectives.}
Differentiable surrogates enable optimization toward
non-differentiable evaluation metrics. Pairwise losses,
including sigmoid approximations, have been studied
for AUC maximization~\cite{yang2023auc}, while
EXACT~\cite{karpukhin2024exact} targets classification accuracy. We use AUC surrogates
to guide online adaptation of training loss
hyperparameters toward task-specific evaluation criteria.

Table~\ref{tab:related_comparison} summarizes existing
approaches by application, loss parameterization,
optimization strategy, and computational complexity.

\section{Method}
\label{sec:method}

We consider the optimization of a training objective
$\mathcal{L}_{\mathrm{tr}}(\theta,w)$, where
$\theta\in\mathbb{R}^{d}$ denotes the model parameters
and $w\in\mathbb{R}^{m}$ is a vector of differentiable
loss hyperparameters. Our goal is to adapt $w$ during
training such that updates of $\theta$ improve a
validation objective $\mathcal{L}_{\mathrm{val}}(\theta)$.
Unlike conventional hyperparameter optimization,
which evaluates multiple training trajectories,
our approach exploits the local dependence of the
training gradient on $w$ to obtain efficient
hyperparameter updates.

\subsection{Core Method}
\label{sec:method_core}

Consider a training step at the current model
parameters $\theta$ and hyperparameters $w_0$.
We first approximate the training objective by
its first-order Taylor expansion with respect
to the hyperparameters:
\begin{equation}
\label{eq:loss_linearization}
\mathcal{L}_{\mathrm{tr}}(\theta,w_0+s)
\approx
\mathcal{L}_{\mathrm{tr}}(\theta,w_0)
+
s^\top \nabla_w
\mathcal{L}_{\mathrm{tr}}(\theta,w_0),
\end{equation}
where $s=w-w_0$ denotes a local hyperparameter
perturbation.

Differentiating Eq.~\eqref{eq:loss_linearization}
with respect to the model parameters yields
\begin{equation}
\label{eq:gradient_linearization}
g(s)\approx g_0+Gs,
\end{equation}
where
\begin{equation}
\label{eq:gradient_jacobian}
\begin{aligned}
g_0 &=
\nabla_\theta
\mathcal{L}_{\mathrm{tr}}(\theta,w_0),\\
G &=
\nabla_\theta\nabla_w^\top
\mathcal{L}_{\mathrm{tr}}(\theta,w_0)
\in\mathbb{R}^{d\times m}.
\end{aligned}
\end{equation}
Thus, the columns of $G$ describe how each loss
hyperparameter modifies the training gradient.
Importantly, Eq.~\eqref{eq:gradient_linearization}
provides a linear approximation to the gradient
even when the original training objective depends
nonlinearly on its hyperparameters.

We now consider a single SGD update,
\begin{equation}
\label{eq:theta_update}
\theta^+(s)=\theta-\eta_\theta g(s),
\end{equation}
where $\eta_\theta$ is the model learning rate.
Let
$v=\nabla_\theta\mathcal{L}_{\mathrm{val}}(\theta)$
denote the validation gradient.
A first-order approximation of the validation
loss after this update gives
\begin{equation}
\label{eq:val_linearization}
\begin{aligned}
\mathcal{L}_{\mathrm{val}}(\theta^+(s))
&\approx
\mathcal{L}_{\mathrm{val}}(\theta)
-\eta_\theta v^\top g(s)\\
&\approx C-\eta_\theta v^\top Gs,
\end{aligned}
\end{equation}
where $C$ is independent of $s$.
Consequently, the local hypergradient is
\begin{equation}
\label{eq:hypergradient}
\nabla_s\mathcal{L}_{\mathrm{val}}
\approx -\eta_\theta G^\top v.
\end{equation}
This leads to a simple hyperparameter update,
\begin{equation}
\label{eq:hyperparameter_update}
w^+
=
w_0+\eta_w G^\top v,
\end{equation}
where $\eta_w$ absorbs the model learning-rate
factor. The update increases the contribution of
hyperparameters whose induced gradient changes
are positively aligned with the validation
gradient.

A direct evaluation of Eq.~\eqref{eq:hypergradient}
would require computing the mixed-derivative
matrix $G$, which can be prohibitively large.
Fortunately, the method only requires its
vector--Jacobian product $G^\top v$.
By the symmetry of mixed derivatives,
\begin{equation}
\label{eq:jvp}
G^\top v
=
D_\theta\left[
\nabla_w\mathcal{L}_{\mathrm{tr}}(\theta,w_0)
\right][v],
\end{equation}
where $D_\theta[\cdot][v]$ denotes a directional
derivative along $v$.

Equation~\eqref{eq:jvp} can be computed using a
Jacobian--vector product (JVP), without explicitly
constructing $G$. Since forward-mode
differentiation propagates a single tangent
direction through the computation, its cost does
not scale linearly with the number of loss
hyperparameters. In practice, this amounts to
one additional forward-mode differentiation
pass through the relevant computation graph,
enabling efficient online hyperparameter
optimization even for high-dimensional $w$.

\subsection{Gradient Similarity}
\label{sec:method_similarity}

The update in Eq.~\eqref{eq:hyperparameter_update}
maximizes the inner product between the training
and validation gradients. Although this improves
their local alignment, the inner product depends
on both gradient direction and magnitude.
Consequently, hyperparameter optimization may
favor increasing the training gradient norm
rather than improving its direction.

This distinction is important because the
magnitude of the training gradient directly
affects the effective model update size.
Even with a fixed learning-rate schedule,
hyperparameter-dependent changes in the gradient
norm implicitly modify the effective learning
rate, potentially interfering with the intended
optimization dynamics.

Since our primary objective is to identify a
training direction that improves validation
performance, we also consider cosine similarity:
\begin{equation}
\label{eq:cosine_objective}
\mathcal{S}(s)
=
\frac{v^\top(g_0+Gs)}
{\|v\|_2\|g_0+Gs\|_2}.
\end{equation}
Unlike the dot product, cosine similarity is
invariant to positive rescaling of the training
gradient and therefore does not directly reward
larger gradient magnitudes.

Differentiating Eq.~\eqref{eq:cosine_objective}
with respect to $s$ gives
\begin{equation}
\label{eq:cosine_gradient}
\nabla_s\mathcal{S}(s)
=
G^\top\left[
\frac{v}{\|v\|_2\|g(s)\|_2}
-
\frac{v^\top g(s)}
{\|v\|_2\|g(s)\|_2^3}g(s)
\right].
\end{equation}
In particular, at $s=0$,
\begin{equation}
\label{eq:cosine_gradient_zero}
\boxed{
\nabla_s\mathcal{S}(0)
=
\frac{G^\top}{\|v\|_2\|g_0\|_2}
\left[
v-
\frac{v^\top g_0}{\|g_0\|_2^2}g_0
\right].
}
\end{equation}

The expression in brackets is the component
of the validation gradient orthogonal to the
current training gradient. Thus, cosine-based
optimization modifies the training direction
without explicitly encouraging changes in
its norm. Both terms in
Eq.~\eqref{eq:cosine_gradient_zero} are
matrix--vector products of the form $G^\top u$
and can therefore be evaluated using the
JVP formulation in Eq.~\eqref{eq:jvp}.

\paragraph{Gradient noise and orthogonalization.}
Despite its desirable scale invariance,
cosine-based optimization introduces an
additional challenge in stochastic training.
The validation gradient and the training gradient
are estimated from different data distributions
and typically exhibit weaker alignment than
the training gradient and its own local
hyperparameter derivatives.

In particular, the correlation between $v$ and
$g_0$ can be small and highly variable across
iterations, while columns of $G$ may remain
strongly correlated with $g_0$. As a result,
the orthogonalization term in
Eq.~\eqref{eq:cosine_gradient_zero} can dominate
the hyperparameter update.

This may produce an undesirable
\emph{unlearning effect}: hyperparameters are
adjusted primarily to suppress components of
the current training gradient, rather than
to increase alignment with the validation
direction. Because stochastic gradients contain
substantial batch-specific variation, such
adjustments may reflect transient gradient
fluctuations instead of systematic differences
between training and validation objectives.

\paragraph{Historical gradient smoothing.}
To mitigate this effect, we replace the
instantaneous gradient used for
orthogonalization with its exponential moving
average (EMA):
\begin{equation}
\label{eq:gradient_ema}
\bar g_t
=
\beta\bar g_{t-1}
+(1-\beta)g_t,
\end{equation}
where $\beta\in[0,1)$ controls the smoothing
strength and $g_t$ denotes the current
training gradient.

The resulting surrogate similarity direction
is defined as
\begin{equation}
\label{eq:smoothed_cosine_direction}
u_t
=
\frac{1}{\|v_t\|_2\|\bar g_t\|_2}
\left[
v_t-
\frac{v_t^\top\bar g_t}
{\|\bar g_t\|_2^2}\bar g_t
\right].
\end{equation}
We then compute the hyperparameter update as
\begin{equation}
\label{eq:smoothed_cosine_update}
w_{t+1}
=
w_t+\eta_w G_t^\top u_t.
\end{equation}

Here, the historical gradient is treated as a
fixed reference direction when constructing
$u_t$. Thus, Eq.~\eqref{eq:smoothed_cosine_update}
is a surrogate hypergradient rather than the
exact derivative of the instantaneous cosine.
The EMA reference suppresses batch-specific
fluctuations and encourages hyperparameter
updates relative to the persistent training
direction, instead of the potentially noisy
current gradient.

Optionally, the validation gradient can also
be smoothed using an EMA to reduce variance
in its estimated direction. The computational
advantage of the method is preserved, since
the final update still requires only a
single product $G_t^\top u_t$.

We investigate both dot-product and
cosine-based objectives experimentally,
analyzing their effects on optimization
stability and downstream performance.

\subsection{Validation Loss}
\label{sec:method_validation}

The proposed method relies on the validation
gradient $v$ to guide training. However,
evaluation metrics are often non-differentiable,
making them unsuitable for direct gradient-based
optimization. Moreover, a conventional
validation loss does not necessarily provide
a locally accurate representation of the
target evaluation metric.

We therefore propose to construct differentiable
surrogate validation objectives that directly
approximate the metric of interest. Unlike
training losses, which may prioritize convenient
optimization properties, these surrogates are
chosen to provide informative local gradient
directions for improving the evaluation metric.

For classification and ranking tasks, we
consider the area under the receiver operating
characteristic curve (ROC AUC). Given validation
scores $a_i^+$ for positive examples and
$a_j^-$ for negative examples, empirical AUC
can be expressed, ignoring ties, as
\begin{equation}
\label{eq:auc}
\mathrm{AUC}
=
\frac{1}{N_+N_-}
\sum_{i=1}^{N_+}
\sum_{j=1}^{N_-}
\mathbb{I}[a_i^+>a_j^-],
\end{equation}
where $N_+$ and $N_-$ denote the numbers
of positive and negative examples. For multiclass classification, we compute the
surrogate independently for each class in a
one-vs-rest setting and average the resulting
losses across classes.

To obtain a differentiable objective,
we replace the pairwise indicator by a
sigmoid approximation:
\begin{equation}
\label{eq:auc_surrogate}
\mathcal{L}_{\mathrm{val}}^{\mathrm{AUC}}
=
\frac{1}{N_+N_-}
\sum_{i=1}^{N_+}
\sum_{j=1}^{N_-}
\sigma\left(
\frac{a_j^- - a_i^+}{\tau}
\right),
\end{equation}
where $\sigma(z)=(1+e^{-z})^{-1}$ and
$\tau>0$ controls the smoothness of the
approximation.

As $\tau\to 0$, the surrogate approaches
the pairwise misranking rate for examples
with unequal scores, corresponding to
$1-\mathrm{AUC}$. For finite $\tau$, it
provides nonzero gradients around the
decision boundary, allowing the method
to identify hyperparameter changes that
improve pairwise ranking.

The validation direction is then obtained as
\begin{equation}
\label{eq:auc_validation_gradient}
v=
\nabla_\theta
\mathcal{L}_{\mathrm{val}}^{\mathrm{AUC}}(\theta).
\end{equation}

By constructing the validation objective
around the actual evaluation criterion,
we aim to improve the local relevance of
the hyperparameter updates. This distinction
is particularly important when the training
objective and the target metric emphasize
different properties of the predictions.

\section{Experiments}
In this section, we evaluate the computational efficiency of the proposed method and its effectiveness in tuning loss hyperparameters across tasks of increasing complexity. We consider a controlled toy example, closed-set classification with shared training and validation classes, and open-set face recognition, where distribution differences introduce additional instability in gradient alignment.

\subsection{Computational Efficiency}
\label{sec:computational_efficiency}

We compare the computational efficiency of the proposed
Jacobian--vector-product (JVP) approach against
two alternative hypergradient computation methods:
reverse-over-reverse automatic differentiation and
implicit differentiation based on a truncated
Neumann series, as employed in
AuxiLearn~\cite{navon2021auxilearn} and
AutoBalance~\cite{li2021autobalance}.
Experiments use a ResNet-50 backbone with ArcFace loss,
a batch size of 32, and 1,000 class-specific margin
hyperparameters on an NVIDIA RTX 4060 GPU.
Implicit differentiation uses five Neumann iterations.
All methods share a precomputed validation gradient,
excluded from the measurements.
We report median computation time and incremental
peak GPU memory consumption.

\begin{figure*}[t]
    \centering
    \includegraphics[width=\textwidth]{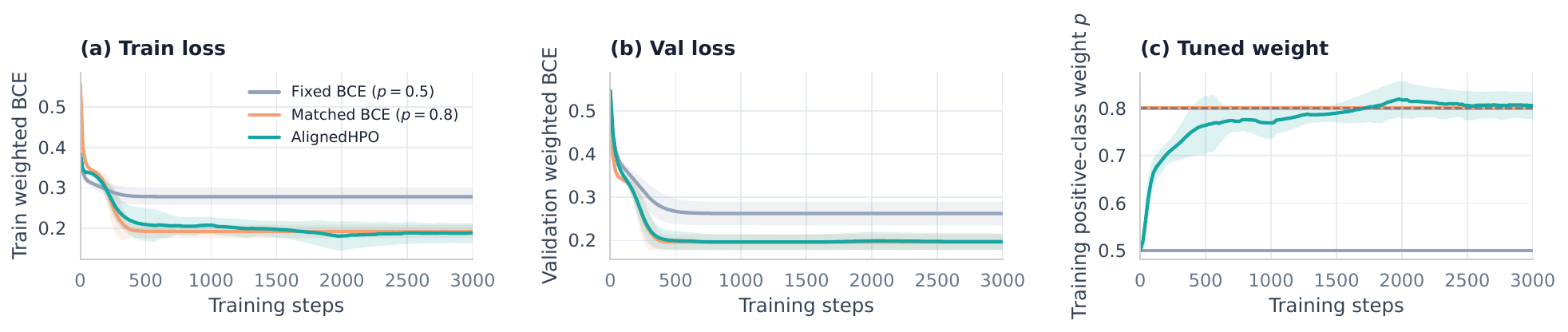}
    \caption{
        Toy example of automatic loss hyperparameter adaptation.
        (a) Training loss evaluated with each method's own
        class weight.
        (b) Validation loss evaluated with the shared weight
        $p_{\mathrm{val}}=0.8$.
        (c) Evolution of the training class weight $p$.
        AlignedHPO automatically adapts the training objective,
        achieving validation performance comparable to the
        oracle baseline with $p=0.8$.
        Solid lines show the mean over five independent runs;
        shaded regions indicate one standard deviation.
    }
    \label{fig:toy_bce_performance}
\end{figure*}

As shown in Fig.~\ref{fig:hypergradient_time},
the proposed JVP approach requires 81.4\,ms,
compared to 162.3\,ms for reverse-over-reverse
and 974.4\,ms for implicit differentiation,
achieving speedups of $1.99\times$ and $11.97\times$,
respectively. JVP and reverse-over-reverse compute
the same local hypergradient, so the improvement
comes solely from the differentiation strategy.

Similarly, Fig.~\ref{fig:hypergradient_memory}
shows that JVP requires 1,703.1\,MiB, compared to
2,061.2\,MiB and 3,055.9\,MiB, reducing memory
consumption by 17.4\% and 44.3\%, respectively.

The higher cost of implicit differentiation
results primarily from repeated Hessian--vector
products used to approximate the inverse Hessian.
Unlike the local methods, it estimates the
sensitivity of an approximately stationary
inner solution and thus computes a different
hypergradient approximation.

Overall, forward-over-reverse differentiation
substantially reduces both runtime and memory
requirements, making it well suited for online
loss hyperparameter optimization.

\subsection{Toy Example: Aligning Training and Validation Objectives}
\label{sec:toy_example}

We consider a one-dimensional binary classification problem
with overlapping class-conditional distributions,
$x \mid y=1 \sim \mathcal{U}(0,2)$ and
$x \mid y=0 \sim \mathcal{U}(0,4)$, with equal class proportions.
The classifier is a linear model $f_\theta(x)=s(x-b)$,
where $s$ and $b$ are learnable parameters.
We optimize the weighted binary cross-entropy loss,
\begin{equation}
\label{eq:toy_bce}
\begin{aligned}
\mathcal{L}(\theta,p) = -\mathbb{E}\big[
    &\,py\log\sigma(f_\theta(x))\\
    &+(1-p)(1-y)\log(1-\sigma(f_\theta(x)))
\big],
\end{aligned}
\end{equation}
where $p$ controls the relative class weights.

We fix the validation weight to $p_{\mathrm{val}}=0.8$
and initialize the training weight at $p_{\mathrm{train}}=0.5$.
We compare standard BCE ($p=0.5$), an oracle baseline
matching the validation objective ($p=0.8$), and
AlignedHPO, which automatically tunes $p$ using gradient alignment.
All methods train the classifier using SGD ($\eta=0.1$);
AlignedHPO additionally optimizes $p$ using Adam
($\eta=0.01$). Results are averaged over five runs
with independently sampled training and validation sets.

Figure~\ref{fig:toy_bce_performance} shows the training
and validation losses and the evolution of $p$.
AlignedHPO reduces the validation loss from approximately
$0.55$ to $0.20$, matching the oracle baseline and
outperforming standard BCE, which converges to approximately
$0.26$. Meanwhile, the learned training weight progressively
increases from $0.5$ toward $0.8$, closely approaching
the validation weight without explicitly targeting its value.
This demonstrates that gradient alignment can automatically
adapt an initially mismatched training objective to recover
oracle-level validation performance.

\subsection{Long-Tailed Classification}
\label{sec:experiments_classification}

We evaluate AlignedHPO on CIFAR-10-LT and CIFAR-100-LT
\cite{cui2019classbalanced,cao2019learning},
constructed by exponentially reducing class frequencies
in the training sets. We use an imbalance ratio of
100 between the most and least frequent classes,
while validation and test sets remain balanced.
This creates a mismatch between the training
distribution and the target evaluation metric.

\paragraph{Parameterized Focal Loss.}
We consider a generalized Focal Loss with
class-dependent weights $\alpha_c$ and focusing
parameters $\gamma_c$:
\begin{equation}
\label{eq:adaptive_focal}
\mathcal{L}_{\mathrm{focal}}
=
-\frac{1}{B}\sum_{i=1}^{B}
\alpha_{y_i}
(1-p_{i,y_i})^{\gamma_{y_i}}
\log p_{i,y_i},
\end{equation}
where $B$ is the batch size and $p_{i,y_i}$
is the predicted probability of the ground-truth
class. We optimize $2C$ hyperparameters, with
class weights normalized using softmax:
\begin{equation}
\label{eq:focal_alpha}
\alpha_c
=
\frac{C\exp(a_c)}
{\sum_{j=1}^{C}\exp(a_j)}.
\end{equation}
The focusing parameters $\gamma_c$ are constrained
to be non-negative.

\paragraph{Compared Methods.}
We compare standard cross-entropy and Focal Loss
with fixed $\gamma=1$ against three adaptive
variants: implicit differentiation (Implicit Diff.)
using a truncated Neumann approximation, and
AlignedHPO with dot-product and cosine-based
gradient alignment.

All adaptive methods optimize the same class-specific
Focal Loss hyperparameters using the differentiable
ROC AUC surrogate introduced in
Sec.~\ref{sec:method_validation}.
We use sigmoid temperatures of $\tau=0.01$ for
CIFAR-10-LT and $\tau=0.001$ for CIFAR-100-LT.
The balanced validation sets guide hyperparameter
optimization, enabling a fair comparison of
different hypergradient estimation strategies.

\paragraph{Training Details.}
We use a ResNet-18~\cite{he2016deep} backbone,
trained for 200 epochs using SGD with momentum 0.9
and an initial learning rate of 0.1.
The learning rate is multiplied by 0.01 at epochs
160 and 180. We use a batch size of 128 for all
methods except implicit differentiation, which
uses 96 due to its higher memory requirements.
The Focal Loss parameters are initialized with
uniform class weights ($\alpha_c=1$) and
zero focusing parameters ($\gamma_c=0$),
corresponding to standard cross-entropy.

\begin{figure*}[t]
    \centering
    \includegraphics[width=\textwidth]{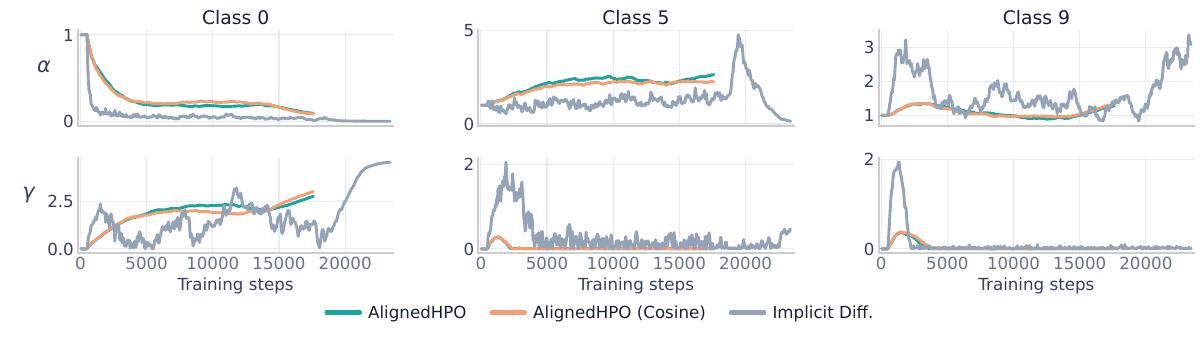}
    \caption{
        Evolution of the effective Focal Loss parameters
        $\alpha$ (top) and $\gamma$ (bottom) for selected
        classes on CIFAR-10-LT. We compare AlignedHPO,
        its cosine variant, and implicit differentiation.
    }
    \label{fig:cifar10_hpo_parameters}
\end{figure*}

\paragraph{Results and Discussion.}
Table~\ref{tab:classification} reports test
classification accuracy on both benchmarks.

In our experiments, all adaptive methods outperform the fixed-loss
baselines, demonstrating the effectiveness of
surrogate-guided hyperparameter optimization.
AlignedHPO improves upon standard cross-entropy
by up to 3.46 and 2.08 percentage points on
CIFAR-10-LT and CIFAR-100-LT, respectively.
It also achieves slightly better results than
implicit differentiation, while offering
substantially lower computational cost
(Sec.~\ref{sec:computational_efficiency}).

\begin{table}[t]
\centering
\caption{Classification accuracy (\%) on CIFAR10-LT and CIFAR100-LT. Higher is better.}
\label{tab:classification}
\setlength{\tabcolsep}{5pt}
\begin{tabular}{lcc}
\toprule
\textbf{Method} & \textbf{CIFAR10-LT} & \textbf{CIFAR100-LT} \\
\midrule
Cross-Entropy & 71.33 & 39.92 \\
Focal ($\gamma=1$) & 72.39 & 39.83 \\
\midrule
Implicit Diff. & 74.53 & 40.48 \\
AlignedHPO (Dot) & \textbf{74.79} & 40.65 \\
AlignedHPO (Cosine) & 74.41 & \textbf{42.00} \\
\bottomrule
\end{tabular}
\end{table}

The two alignment objectives exhibit different
behavior. Dot-product alignment performs slightly
better on CIFAR-10-LT (74.79\% vs. 74.41\%),
whereas cosine alignment achieves a larger
improvement on CIFAR-100-LT (42.00\% vs. 40.65\%).
This suggests that controlling gradient magnitude
may be more beneficial in more challenging
classification settings.

Figure~\ref{fig:cifar10_hpo_parameters} illustrates
the evolution of learned hyperparameters on
CIFAR-10-LT. We expect larger class weights
$\alpha_c$ for rare classes to compensate for
imbalance, and larger focusing parameters
$\gamma_c$ for frequent classes to suppress
their easier examples. The learned parameters
partially follow this intuition: $\alpha_0$
decreases substantially, while $\gamma_0$
increases, reducing the contribution of the
most frequent class. Meanwhile, $\alpha_5$
increases, although the rarest class retains
a weight close to its initialization.
The focusing parameters for classes 5 and 9
remain close to zero.

Notably, both AlignedHPO variants produce
smooth and similar hyperparameter trajectories,
whereas implicit differentiation exhibits
substantially larger fluctuations and occasional
sharp changes. This indicates more stable
hyperparameter adaptation with the proposed
local gradient alignment approach, despite its
considerably lower computational cost.

\subsection{Open-Set Face Recognition}
\label{sec:experiments_face}

We further evaluate AlignedHPO on open-set face
recognition, where training and evaluation identities
are disjoint. Unlike closed-set classification,
this introduces an additional mismatch between
training and validation objectives, making
hyperparameter optimization more challenging.

We train models on CASIA-WebFace~\cite{yi2014learning} and evaluate
verification accuracy on LFW~\cite{huang2007lfw}, AgeDB-30~\cite{moschoglou2017agedb}, and CFP-FP~\cite{sengupta2016frontal},
which capture variations in imaging conditions,
age, and pose, respectively.

\paragraph{Parameterized ArcFace Loss.}
We adopt ArcFace as the training objective,
which introduces an additive angular margin
into normalized softmax classification:
\begin{equation}
\label{eq:arcface_loss}
\mathcal{L}_{\mathrm{ArcFace}}
=
-\frac{1}{B}\sum_{i=1}^{B}
\log
\frac{e^{s\cos(\theta_{i,y_i}+m)}}
{e^{s\cos(\theta_{i,y_i}+m)}
+\sum_{j\ne y_i}e^{s\cos\theta_{i,j}}},
\end{equation}
where $\theta_{i,j}$ denotes the angle between
the normalized embedding and class weight,
$m$ is the angular margin, and $s$ is the
logit scaling factor. We treat these parameters
as differentiable hyperparameters and optimize
them during training.

\paragraph{Validation Objective.}
Face verification relies on similarity scores
between image pairs rather than class predictions.
We compute cosine similarity between normalized
embeddings and use the differentiable binary
ROC AUC surrogate from
Sec.~\ref{sec:method_validation}
to guide hyperparameter optimization.
This allows the classification-based training
loss to be adapted toward the verification
objective.

\paragraph{Dataset-Specific Optimization.}
We consider three variants of AlignedHPO,
independently tuned using validation pairs
from LFW, AgeDB-30, and CFP-FP.
All variants share the same CASIA-WebFace
training data and differ only in the
validation distribution used for tuning.
We evaluate both dot-product and cosine-based
alignment against standard ArcFace with
fixed hyperparameters ($s=32, m=0.5$).

\paragraph{Training Details.}
We use a MobileFaceNet backbone~\cite{chen2018mobilefacenets} trained for
18 epochs with SGD, Nesterov momentum of 0.9,
and weight decay of $5\times10^{-4}$.
The initial learning rate is 0.01 and is
multiplied by 0.1 at epochs 6, 11, and 16.

Since the ArcFace scale parameter strongly
affects the training gradient magnitude,
we normalize the gradients using an exponential
moving average (EMA) of their norm, with a
smoothing factor of 0.9. This reduces variations
in the effective learning rate caused by
hyperparameter adaptation and ensures more
comparable optimization dynamics across methods.

\paragraph{Results and Discussion.}
Table~\ref{tab:face_recognition} reports
verification accuracy on all three benchmarks.

\begin{table}[t]
\centering
\caption{Face verification performance (\%) on LFW, AgeDB-30, and CFP-FP. Each AlignedHPO variant is tuned using the corresponding validation dataset. The retrained variants use the parameters identified during hyperparameter optimization.}
\label{tab:face_recognition}
\setlength{\tabcolsep}{4pt}
\begin{tabular}{lccc}
\toprule
\textbf{Method} & \textbf{LFW} & \textbf{AgeDB} & \textbf{CFP-FP} \\
\midrule
ArcFace (original) & \textbf{97.68} & \textbf{86.37} & \underline{87.34} \\
\midrule
\multicolumn{4}{l}{\textit{AlignedHPO -- Dot product}} \\
AlignedHPO (LFW) & \underline{97.15} & 84.33 & 86.63 \\
AlignedHPO (AgeDB) & 96.98 & \underline{85.87} & 79.44 \\
AlignedHPO (CFP-FP) & 96.68 & 82.45 & \textbf{89.26} \\
\midrule
\multicolumn{4}{l}{\textit{AlignedHPO -- Cosine similarity}} \\
AlignedHPO (LFW) & 96.73 & 82.40 & 84.01 \\
AlignedHPO (AgeDB) & 97.13 & 85.45 & 81.97 \\
AlignedHPO (CFP-FP) & 96.68 & 81.58 & 88.17 \\
\bottomrule
\end{tabular}
\end{table}

The results demonstrate dataset-specific adaptation
of the learned loss hyperparameters. For both
alignment objectives, tuning on CFP-FP achieves
the highest CFP-FP accuracy among adaptive variants,
while tuning on AgeDB-30 provides the best
AgeDB-30 performance. A similar trend is observed
on LFW with dot-product alignment. This indicates
that AlignedHPO can adapt the training objective
to the characteristics of the target validation
distribution, even when training and validation
identities are disjoint.

However, the original ArcFace baseline remains
superior on LFW and AgeDB-30. The main improvement
is observed on CFP-FP, where dot-product alignment
achieves 89.26\%, exceeding the baseline by
1.92 percentage points. These results suggest
that optimizing hyperparameters for a single
validation distribution can improve performance
on that target but may reduce generalization
to other verification benchmarks.

Another possible limitation is the mismatch
between the validation objective and evaluation
metric: we optimize a differentiable ROC AUC
surrogate but evaluate verification accuracy.
Although related, these metrics are not equivalent,
potentially limiting the effectiveness of
hyperparameter adaptation.

Future work could address these limitations by
jointly optimizing hyperparameters across multiple
validation datasets and exploring differentiable
accuracy surrogates~\cite{karpukhin2024exact}.
These extensions may improve the robustness
and generalization of the learned training objective.

% \midrule
% \multicolumn{4}{l}{\textit{AlignedHPO -- Retrained with tuned parameters}} \\
% AlignedHPO (LFW) & 97.23 & 83.02 & 85.36 \\
% AlignedHPO (AgeDB) & 96.98 & 85.13 & 81.03 \\
% AlignedHPO (CFP-FP) & 97.07 & 82.77 & \textbf{89.50} \\

\section{Limitations}
\label{sec:limitations}

AlignedHPO has several limitations. First, its
local, single-step approximation neglects the
long-term effects of hyperparameters on training,
potentially limiting effectiveness when delayed
effects are important.

Second, the method depends on validation gradient
quality. Differences between training and validation
distributions can introduce noise and instability.
Although gradient smoothing mitigates these effects,
our open-set experiments show that dataset-specific
adaptation does not necessarily generalize across
evaluation benchmarks.

Finally, performance depends on how well the
differentiable validation objective approximates
the target metric. In particular, our ROC AUC
surrogate does not directly optimize verification
accuracy. Future work could explore multi-dataset
validation objectives, alternative metric surrogates,
and extensions beyond local hypergradient estimation.

\section{Conclusion}
\label{sec:conclusion}

We introduced AlignedHPO, an efficient framework
for online optimization of differentiable loss
hyperparameters. By combining local gradient
alignment with Jacobian--vector products,
the proposed method avoids expensive
trajectory differentiation and inverse-Hessian
approximations while supporting nonlinear
loss parameterizations.

We further investigated dot-product and
cosine-based alignment and introduced gradient
smoothing to improve optimization stability.
Experiments demonstrate substantial reductions
in hypergradient computation time and memory,
along with competitive or improved performance
on long-tailed classification and open-set face
recognition. Our results also highlight the
potential of differentiable metric surrogates
for guiding loss hyperparameter optimization,
providing a practical foundation for automated
loss adaptation during deep network training.

\section*{LLM Usage Statement}
OpenAI ChatGPT was used to assist with scientific
writing and editing, related-work analysis, code
development, and experimental plotting.
All technical contributions, experimental results,
and conclusions were developed and verified
by the authors.

{
    \small
    \bibliographystyle{ieeenat_fullname}
    \bibliography{main}
}


\end{document}